\documentclass[letterpaper, 10 pt, conference]{ieeeconf}  

\IEEEoverridecommandlockouts                              

\usepackage{amssymb}
\usepackage{xcolor}
\definecolor{idcolor}{HTML}{A6CE39}
\usepackage{graphicx}
\graphicspath{{figures/}}
\usepackage{array}
\usepackage{amsmath}
\usepackage{tikz}

\usepackage{scalerel}
\usetikzlibrary{svg.path}
\definecolor{orcidlogocol}{HTML}{A6CE39}
\tikzset{
  orcidlogo/.pic={
    \fill[orcidlogocol] svg{M256,128c0,70.7-57.3,128-128,128C57.3,256,0,198.7,0,128C0,57.3,57.3,0,128,0C198.7,0,256,57.3,256,128z};
    \fill[white] svg{M86.3,186.2H70.9V79.1h15.4v48.4V186.2z}
                 svg{M108.9,79.1h41.6c39.6,0,57,28.3,57,53.6c0,27.5-21.5,53.6-56.8,53.6h-41.8V79.1z M124.3,172.4h24.5c34.9,0,42.9-26.5,42.9-39.7c0-21.5-13.7-39.7-43.7-39.7h-23.7V172.4z}
                 svg{M88.7,56.8c0,5.5-4.5,10.1-10.1,10.1c-5.6,0-10.1-4.6-10.1-10.1c0-5.6,4.5-10.1,10.1-10.1C84.2,46.7,88.7,51.3,88.7,56.8z};
  }
}

\newcommand\orcidicon[1]{\href{https://orcid.org/#1}{\mbox{\scalerel*{
\begin{tikzpicture}[yscale=-1,transform shape]
\pic{orcidlogo};
\end{tikzpicture}
}{|}}}}

\makeatletter
\newcommand*{\emails}[2][@in.tum.de]{%
    \def\@tempa{\@gobble}%
    \@for\qrr@email:=#2\do{%
        \edef\@tempb{\noexpand\href{mailto:\qrr@email #1}{\qrr@email}}%
        \edef\@tempa{\unexpanded\expandafter{\@tempa}{, }\unexpanded\expandafter{\@tempb}}}%
    \{\@tempa\}#1%
}

\makeatletter
\let\NAT@parse\undefined
\makeatother
\usepackage{hyperref} 
\hypersetup{
    bookmarks=false,
    colorlinks=true,
    citecolor=green,
    linkcolor=blue,
    filecolor=magenta,      
    hypertexnames=true,
    urlcolor=cyan,
    pagebackref=true,
    pdftitle={ITSC2022},
    pdfpagemode=FullScreen
    }
\title{\LARGE \bf
A Survey of Robust LiDAR-based 3D Object Detection Methods for Autonomous Driving}
\author{Walter~Zimmer\orcidicon{0000-0003-4565-1272}, Eme\c{c} Er\c{c}elik\orcidicon{0000-0002-0716-0475}, Xingcheng Zhou\orcidicon{0000-0003-1178-5221}, \\Xavier Jair Diaz Ortiz\orcidicon{0000-0003-0918-913X}, and Alois~Knoll\orcidicon{0000-0003-4840-076X},~\IEEEmembership{Life~Fellow,~IEEE}
\thanks{The authors are with the Department of Informatics, Technical University of Munich (TUM), Garching, Germany.\protect\\
E-mail: \tt\small \href{mailto:walter.zimmer@cs.tum.edu}{walter.zimmer@cs.tum.edu}, 
\emails{ercelik,zhou,diaz,knoll}
}
}

\begin{document}

\maketitle
\thispagestyle{empty}
\pagestyle{empty}

\begin{abstract}
The purpose of this work is to review the \textit{state-of-the-art} \textit{LiDAR}-based 3D object detection methods, datasets, and challenges. We describe novel data augmentation methods, sampling strategies, activation functions, attention mechanisms, and regularization methods. Furthermore, we list recently introduced normalization methods, 
learning rate schedules and loss functions.
Moreover, we also cover advantages and limitations of 10 novel autonomous driving datasets. We evaluate novel 3D object detectors on the \textit{KITTI}, \textit{nuScenes}, and \textit{Waymo} dataset and show their accuracy, speed, and robustness. Finally, we mention the current challenges in 3D object detection in \textit{LiDAR} point clouds and list some open issues.
\end{abstract}


\section{Introduction}\label{sec:introduction}
Autonomous driving (AD) is increasingly gaining attention worldwide and can lead to many advantages to the population. The potential of this technology is clear, and it is predicted that it will dramatically change the  transportation sector. The application of robust 3D Object Detection in autonomous driving is vital. In this context, robustness means to cope with detection errors (e.g., occlusions) and erroneous input (e.g., sensor noise). Sometimes, the environment is not optimal for detecting objects (e.g., in rain, snow, fog or bright sunlight). Robust detection stands for a good generalization to detect unseen objects in a different environment. 

Besides, infrastructure sensors (like the ones in the \textit{Providentia} \cite{krammer2019providentia} system) support the perception of the environment, improve traffic safety, and offer higher robustness and performance through different mounting positions. Multiple view points help to better detect objects in 3D (position, orientation, and size), and to increase the robustness against sunlight, snow, dust, and fog. 

The aim of this survey paper is to provide an overview of novel 3D object detection methods and tricks. 
The best LiDAR-based 3D object detection algorithm on the KITTI dataset is 68.63\% more accurate (on the 3D car class) than the best camera-only 3D object detection algorithm on that dataset. At night, the LiDAR performs better and is able to detect objects within a range of 50-60 m (Ouster OS1-64 gen. 2 LiDAR).


Our contribution is partitioned into the following:
\begin{itemize}
    \item We provide \textit{state-of-the-art} LiDAR-based 3D object detectors and show their accuracy, speed, and robustness.
    \item Furthermore, we describe novel data augmentation methods, sampling strategies, activation functions, attention mechanisms, and regularization methods.
    \item We list recently introduced normalization methods, 
    learning rate schedules and loss functions.
    \item We compare the most important datasets in detail, list their advantages and limitations, and evaluate the \textit{state-of-the-art} detectors on the \textit{KITTI} \cite{GeigerVisionMeetsRobotics2013}, \textit{nuScenes} \cite{caesarNuScenesMultimodalDataset2020}, and \textit{Waymo} \cite{sun2020scalability} datasets in terms of \textit{mean average precision} (\texttt{mAP}) and inference speed. 
    \item Finally, we mention current challenges in 3D object detection in LiDAR point clouds, list some open issues and provide research directions for the future.
\end{itemize}
\section{Related Work}
\cite{wuDeep3DObject2021} provides a fundamental analysis of 3D object detection stages. They compare several object detection network architectures, 3D box encodings, and evaluation metrics. \cite{guoDeepLearning3D2020} covers deep learning methods for several point cloud understanding tasks, including 3D object detection. \cite{arnoldSurvey3DObject2019} provides an overview of available datasets and evaluates the performance of object detection models. The authors summarize 3D object detection advancements for autonomous driving vehicles. Lastly, research gaps and future research directions are presented.

\cite{huangSurveyStateofArtAutonomous2020} separates \textit{LiDAR}-based 3D object detection methods into 3D volume-based and projection-based methods, which consist of \textit{birds-eye-view} (\texttt{BEV}) and  \textit{frontal view} (\texttt{FV}) projections. Furthermore, the authors list nine datasets for 3D object detection and autonomous driving. \cite{rahmanNoticeViolationIEEE2020} describes the basic concepts of 3D bounding box encoding techniques, 3D data representation and sensing modalities. The authors introduce datasets that are helpful for deep learning-based 3D object detection projects. Furthermore, an in-depth systematic review of recent 3D object detection methods is shown as well as their limitations. 
\section{State-of-the-Art}
\label{cha:state_of_the_art}

\begin{figure}[htbp]
    \centering
    \includegraphics[width=1.0\linewidth]{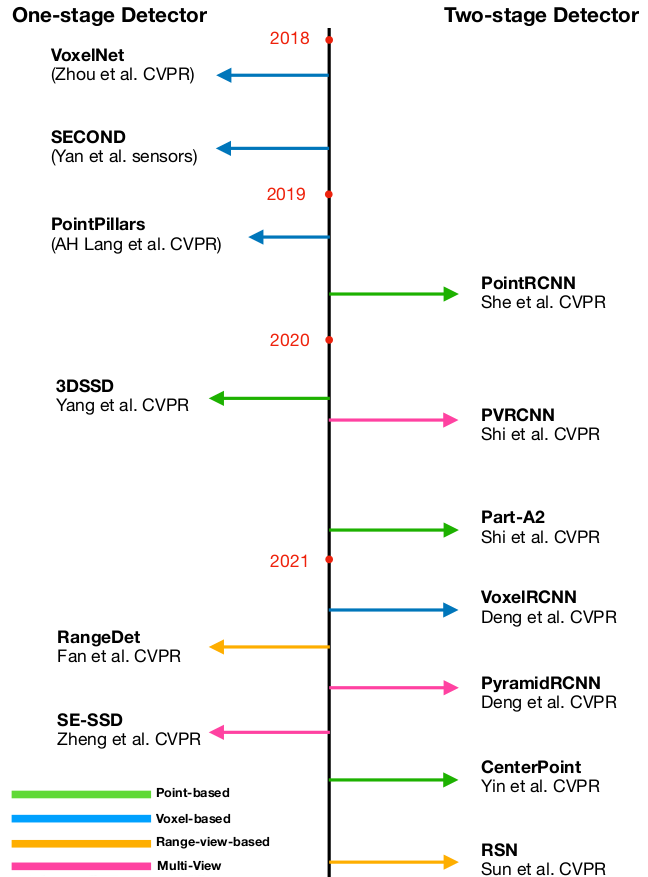}
    \caption{Overview of the state-of-the-art detectors.}
    \label{fig:overview_sota}
\end{figure}

3D Object Detection within point cloud data can be partitioned into a) point-based (input-wise permutation invariant) methods, b) voxel-based (grid-representation-based) methods, c) range-view-based methods and d) multi-view-based methods. Figure \ref{fig:overview_sota} shows a chronological view of the state-of-the-art detectors.

\subsection{3D Object Detection Architectures}
We group novel architectures into one-stage architectures (\textit{PointPillars}, 
\textit{3DSSD}, 
\textit{SA-SSD}, 
\textit{CIA-SSD} and \textit{SE-SSD}) and two-stage architectures (\textit{PV-RCNN}, \textit{PV-RCNN++}, and \textit{CenterPoint} and list their advantages, limitations and possible improvements. \\

\textbf{PointPillars}. This architecture is a point and voxel based method. First, the point cloud is divided into grids in the x-y coordinates, creating a set of pillars. Each point in the cloud (x, y, z, reflectance) is converted from a 4D vector into a 9D vector. The points are extended with the distance to the arithmetic mean of the pillar and the distance to the center of the pillar. Internally, \textit{PointNet} is used to extract dense robust features from sparse \textit{pillars}. Each 2D voxel cell (\textit{pillar}) is represented by a fixed-size feature vector. 2D voxelization grids are more computationally efficient than those detection networks that take 3D voxelization grids as input. The 2D voxelization representation doesn't suffer from scale ambiguity and occlusion \cite{sanatkarLidar3dObject2020}. The encoder learns the point cloud representations from the \textit{pillars}. In the end, a 2D CNN (modified SSD) is applied for joint bounding box and classification prediction. The end-to-end trainable \textit{PointPillars} network runs at 62 Hz using \textit{TensorRT}. \textit{TensorRT} is a library for optimized GPU inference. \cite{langPointPillarsFastEncoders2019} shows that switching to \textit{TensorRT} increases inference speedup from 42.4 Hz to 62 Hz). It is highly efficient, because all key operations can be formulated as 2D convolutions that are extremely efficient to compute on a GPU. One limitation is that sometimes pedestrians and cyclists are misclassified. Pedestrians are confused with narrow vertical features of the environment such as poles or tree trunks. Cyclist are often misclassified because they are not much represented in the dataset.\\

\textbf{3DSSD}. \textit{3DSSD} \cite{yang20203dssd} (a single-stage 3D detection architecture) optimizes the inference speed while keeping the accuracy of two-stage detectors without voxel usage. The raw point-based 3D detectors first sample key points and extract features using set abstraction modules. To cover all objects in the scene, it's necessary to obtain raw point features from the extracted features of the key points. Since this process slows down the inference, the authors propose a new point sampling method, that uses the feature distance metric for key point sampling in addition to the Euclidean distance. This helps to increase the recall of the points and to balance positive and negative point samples. They also add a shift to the sampled candidate points before regression to have a better center prediction since the network is anchor-free. Additionally, the ground truth classification scores are changed with a value related to the distance of the candidate points to the surfaces of the ground truth bounding box, and therefore provides more realistic goals and further enhances the AP results. With all the extensions, the proposed architecture can reach an inference speed of 25 Hz. One downside of 3DSSD is its single-stage architecture that limits the 3D AP to 79.57\% for the Car class (moderate difficulty).\\

\textbf{SA-SSD}. \textit{SA-SSD} \cite{he2020structure} (a single-stage 3D object detector) utilizes 3D convolutions on voxels as a backbone and 2D convolutions for the refinement and regression of the final 3D bounding boxes. The contribution of the architecture mainly arises from the proposed auxiliary network, that applies foreground segmentation and center estimation tasks on the interpolated LiDAR points and features. An additional segmentation task increases the sensitivity of the backbone to the boundaries of the objects. The center estimation enhances the relation of object points by learning to estimate object-specific centers from point-specific features. They also propose a feature map warping module to relieve the inconsistency between the positions of the refined anchors\footnote{Anchor boxes are essential if the input suffers from scale ambiguity, like natural images where the dimensions of the objects appearing in images depend on their distances to the camera, or when the network is required to detect different class objects (cars, bicycles and pedestrians) with different dimensions.} and the classification scores by calculating classification as a consensus of object-related regions from the regression branch. With the given extensions and the ability of taking the auxiliary network out during inference, the \textit{SA-SSD} network reaches 25 Hz, while improving the best scores of the single-stage detectors and reaching the accuracy of two-stage detectors.  \\

\textbf{CIA-SSD}. \cite{Zheng2021} proposed a new single-stage detector. Their contribution is a light-weight spatial-semantic \textit{feature aggregation} module, that adaptively fuses high-level abstract semantic features and low-level spatial features for more accurate bounding box regression and classification. An \textit{IoU-aware} confidence rectification module alleviates the misalignment between the localization accuracy and classification confidence. A distance-variant \textit{IoU-weighted} \texttt{NMS} smooths the regressions and avoids redundant predictions.\\

\textbf{PV-RCNN}. Recently, \cite{shiPVRCNNPointVoxelFeature2020} introduced the \textit{PV-RCNN} architecture that integrates two point cloud feature learning strategies, a voxel-based and \textit{PointNet}-based network. The goal is to use the best methods from both of them. First, a small set of key points is sampled from the point cloud to encode the whole scene. In parallel, it creates voxels from the raw point cloud and then encodes these 3D voxels to key points. The proposed \textit{Voxel Set Abstraction} module incorporates the multi-scale voxel features into the nearby key points with the predefined balls and radii. The result is a key point feature vector containing information about the sampled key point and the 3D sparse convolutions. The key point features are reweighted by the \textit{Predicted key point Weighting} (PKW) module. The motivation behind the reweighting is that foreground key points should contribute more to the accurate refinement of the proposals than the background key points. To group the key point features for proposal refinement, the authors propose the \textit{RoI-grid pooling} via \textit{Set Abstraction} (SA). Here, the key point features are aggregated (abstracted) to \textit{RoI} grid points. In the end, a 3D bounding box refinement is taking place. \\

\textbf{PV-RCNN++}. \textit{PV-RCNN++} \cite{shi2021pv} (a two-stage detection network) further improves the accuracy (+4\% \texttt{mAP} on \textit{Waymo}) and inference speed (+6 FPS\footnote{On the \textit{KITTI} dataset, the inference speed has improved from 13 to 17 FPS (+4 FPS).} on \textit{Waymo}). The authors proposed a new sampling strategy (\textit{sectorized proposal-centric} strategy) to produce representative key points. It concentrates the limited key points to be around 3D proposals to encode more effective features for scene encoding and proposal refinement. Furthermore, a novel local feature aggregation method (\textit{VectorPool} aggregation) was introduced. It replaces the previous set abstraction (SA) method by aggregating local point cloud features. The new method consumes less resources, and can handle large number of points.\\

\textbf{CenterPoint} \cite{yin2021center}. In a first stage, the centers of objects are found, from which other attributes such as size, orientation, and velocity are estimated by different regression heads. By considering 3D objects as rotational-invariant points, this network is more flexible in the prediction of rotated objects with higher accuracy than anchor-based detectors. In a second stage, other point features, coming from the face centers of the initial bounding box estimate, are condensed into a vector and fed into a \texttt{MLP} for prediction of confidence scores as well as for further refinement of the box attributes. The detection head is implemented as a heatmap in BEV. In this manner, every local peak of such heatmap corresponds to the center of a detected object. After processing the input point cloud for feature extraction, the result is a map-view, that is passed to a 2D CNN detection head to produce a K-channel heatmap -- one channel for each K class respectively -- with peaks at the center location of the detected object. In the training phase, the centers of ground truth bounding boxes are projected into a map-view so that the detection head can target a 2D Gaussian around these projections using a focal loss. A trained model on \textit{nuScenes} with a \textit{PointPillars} backbone runs at 31 FPS. \\

\textbf{SE-SSD.} \cite{zheng2021se} proposes a one-stage approach and adopts a single-stage detection architecture consisting of a teacher \& student model. Both models share the same structure that resembles the CIA-SSD \cite{Zheng2021} architecture. The teacher model makes predictions based on the original point cloud, these are transformed into "soft targets" used to train the student. The student model makes predictions based on especially augmented data and is trained on the hard and soft targets via a \textit{Consistency Loss} as well as a new \textit{Orientation Aware Distance-IoU} loss (ODIoU). This loss emphasizes the orientation of the bounding boxes as well as the alignment of the center points. The authors propose a new \textit{Shape-Aware Data Augmentation} scheme that performs dropout, swapping and sparsification of points. This data augmentation is applied to the point cloud data the student model is trained on. In this way, by using both a teacher and a student single-stage object detector, the framework can boost the precision of the detector significantly without incurring extra computation during the inference. At the same time, it also inherits the advantages of one-stage methods, namely the speed.

\subsection{Data Augmentation}
The goal of data augmentation is to improve generalization and make networks to become invariant with respect to rotation, translation, and natural changes in point clouds. Randomness of these data augmentation methods increases the generalization ability. Recently, it has become common to apply these data augmentation methods individually to bounding boxes of objects, which are called object-level data augmentation methods. Some common point cloud data augmentations are rotation around the vertical z-axis, flipping with respect to x-z and y-z planes, sparsification, additive \textit{Gaussian noise}, \textit{frustum dropout}, and scene augmentation with object point cloud frames. \textit{AutoAugment} \cite{hataya2020faster} formulates the sequential discrete decision problem as a \textit{Reinforcement Learning} (RL) problem. The goal is to learn data augmentation strategies for object detection. It is shown that using the data augmentation policies generated by the RL agent outperforms the random data augmentation approach. Most of these data augmentation methods can either be applied at the scene-level or at object-level. \cite{Improving3DObject2020} proposed a method that automates the design of data augmentation policies using an evolutionary search algorithm. Recently, \textit{PointCutMix} \cite{zhang2021pointcutmix} was proposed that finds the optimal assignment between two point clouds and generates new training data by replacing the points in one sample with their optimal assigned pairs.

\subsection{Sampling Strategies}
\textit{Farthest-Point-Sampling} (FPS) \cite{eldar1997farthest} samples key points in different sectors to keep the uniformly distributed property of key points while accelerating the key point sampling process. It is not the optimal strategy because of quadratic complexity.
The \textit{sectorized proposal-centric key point sampling strategy} proposed in \cite{shi2021pv} concentrates the limited key points to be around 3D proposals to encode more effective features for scene encoding and proposal refinement. This strategy is much more efficient and can handle large-scale 3D scenes with million of points efficiently.

\subsection{Activation Functions}
The choice of activation functions is unavoidable when we design neural networks. They are introduced to ensure the non-linearity of the network, because the expressive power of linear models is usually insufficient, and can not satisfy the needs of various tasks in artificial intelligence. In neural networks, we usually add activation functions after each affine transformation to project and then keep activated features. \textit{Swish} \cite{Ramachandran} was proposed in 2017, its definition is $Swish(x) = x*Sigmoid(x)$. The curve of \textit{Swish} is similar to \textit{ReLU}, but its derivative is continuous at 0 and its performance is slightly better. In the meanwhile, it brings higher computation cost. To reduce the computation cost, \textit{Hard Swish} \cite{Zhang2019} uses \textit{ReLU6} to approximate the \textit{Swish} function. Its definition is $HardSwish(x)=x\frac{ReLU6(x+3)}{6}$. In comparison to \textit{Swish}, its computation cost is much cheaper, so that it recently is getting high popularity.

\subsection{Attention Mechanisms}
\textit{Attention mechanisms} are specially-designed parts of neural networks that learn to put emphasis on particularly important regions in the data. 3D Object Detection is one of the tasks where these mechanisms have recently been applied to. Especially, \textit{attention mechanisms} can be considered to strengthen the robustness of voxel feature learning. \textit{TANet} \cite{liu2020tanet} presents a \textit{triple attention} module embedded in the voxel feature extraction process and combines it with a proposed \textit{cascaded refinement network}. Li \emph{et~al.} \cite{li20203d} implemented a novel feature pooling module (\textit{ACA module}), to obtain a \textit{perspective-invariant} prediction head. The attention mechanism is introduced to weight the contribution of each  perspective differently. Existing \texttt{IoU} prediction methods suffer from the perspective variance. The proposed \textit{ACA module} aggregates a local point cloud feature from  each perspective of eight corners in the bounding box and adaptively weights the contribution of each perspective with a novel \textit{perspective-wise} and \textit{channel-wise} attention mechanism. Compared to the models that treat each corner and channel equally, the performance (\texttt{mAP}) increases slightly when the extracted features are re-weighted by the \textit{perspective-wise} and \textit{channel-wise} attention.
\textit{CenterNet3D} \cite{wang2020centernet3d} proposed an auxiliary \textit{corner attention module} to enforce the CNN backbone to pay more attention to object boundaries, and to learn discriminative corner features. This leads to more accurate 3D bounding boxes. The corner attention module is implemented with a \textit{corner classification module} that leads to a higher performance (+1-2\% AP) without extra cost.

\subsection{Regularization Strategies}
Regularization strategies are designed to reduce the test error of a machine learning algorithm, possible at the expense of increased training error. While \textit{Dropout} \cite{srivastava2014dropout} is the most famous regularization method in the field of deep learning, many new methods were proposed recently. \textit{DropBlock} \cite{ghiasi2018dropblock} is a structured form of \textit{Dropout} directed at regularizing convolutional networks. In \textit{DropBlock}, units in a continuous region of a feature map are dropped together. As \textit{DropBlock} discards features in a correlated area, the networks must look elsewhere for evidence to fit the data. The idea is to prevent co-adaption, where the neural network becomes too reliant on particular areas, as this could be a symptom of overfitting.

\subsection{Normalization Methods}
The normalization method can help to generalize neural networks and accelerate the training process. \textit{Batch Normalization} \cite{Ioffe2015} was proposed to solve the \textit{internal covariate shift} problem, which means the distribution of the layer inputs may get greatly changed after several weight updating iterations during the training process. It scales the output of the layer by standardizing the activated output of a node per mini-batch, and the standardized output is subject to the distribution with a mean of zero and a standard deviation of one. It also reduces the model dependence on the initial values of the parameters. \textit{Filter Response Normalization} \cite{Singh} is a combination of normalization and activation function. It performs normalization on each activation channel of each batch separately, during which the dependence of other batches is eliminated. \textit{Attentive Normalization} \cite{Li} introduces the \textit{attention mechanism} to the normalization technique. It re-weights the channel-wise affine transformation component by learning the instance-specific weights and computing the weighted sum of mixture for feature re-calibration.


\subsection{Learning Rate Schedules}
At the beginning of training, the weights of a model are randomly initialized where the model may become unstable with a large learning rate. To solve this problem, \textit{Warmup} \cite{Loshchilov} chooses a relatively small learning rate at the beginning of the training process and then increases it a few epochs later. \textit{Cosine Annealing} is widely applied in the training of 3D object detection models such as in \textit{PV-RCNN} \cite{Shi} and \textit{CIA-SSD} \cite{Zheng2021}. It's a straightforward strategy that the learning rate decreases following a cosine function where the value declines slowly at the beginning and the end while drops rapidly in the middle. The property of cosine functions fits well with the learning rate and sometimes it is also combined with warm restart. In warm restart, cosine annealing is repeated. The restarted learning rate depends on the parameters of the last iteration instead of a fixed value. \textit{Warmup} and the cosine annealing strategy can also be combined \cite{loshchilovSGDRStochasticGradient2017}. 

\subsection{Loss Functions}
The default classification loss for object detection networks is the \textit{cross-entropy loss} that is equal to the negative log likelihood (NLL) of the ground truth class. The main complication of the classification loss of object detection networks is to address the class imbalance between the background (negative) class and positive classes. Most of the potential bounding box candidates contain background scenes and not target objects. The two standard approaches to address this class imbalance are \textit{hard negative mining}
and \textit{focal loss} \cite{lin2017focal}. In the \textit{hard negative mining}, negative bounding boxes are sub-sampled per scene such that the number of negative boxes is at most three times higher. \textit{Focal loss} adaptively adjusts the contributing weight of each example in the classification loss during the training process instead of explicitly sub-sampling the negative examples. 

The \textit{PointPillars} architecture combines three loss functions: localization loss ($\mathcal{L}_{loc}$), classification loss ($\mathcal{L}_{cls}$) and the direction loss ($\mathcal{L}_{dir}$).
\begin{align*}
\mathcal{L}_{loc}  &=\sum_{b\in (x,y,z,w,l,h,\theta)}{SmoothL1(\Delta b)}  \\
\mathcal{L}_{cls}  &=-\alpha_a(1-p^a)^\gamma log(p^a) \\
\mathcal{L}_{dir}  &= SmoothL1(sin(\theta_p-\theta_t))\\
\mathcal{L}_{total}&=\frac{1}{N_{pos}}(\beta_{loc}\mathcal{L}_{loc}+\beta_{cls}\mathcal{L}_{cls}+\beta_{dir}\mathcal{L}_{dir})
\end{align*}

\textit{Focal loss} is used as object classification loss ($\mathcal{L}_{cls}$), and softmax classification loss is used as direction loss ($\mathcal{L}_{dir}$). The total loss is a weighted sum of all three losses.

\cite{zhou2019iou} proposed an \textit{IoU loss} for rotated 3D bounding boxes (that considers the rotation in two directions), since an axis-aligned box is not suitable for representing target objects in 3D. This loss can be applied for both axis-aligned or rotated 2D/3D objects. The \texttt{IoU} between two 3D bounding boxes (d and g) that are rotated in one direction (e.g. yaw) can be calculated as follows: 
\begin{align*}
    IoU_{3D} = \frac{Area_{overlap} \cdot h_{overlap}}{Area_g \cdot h_g + Area_d \cdot h_d - Area_{overlap} \cdot h_{overlap}}
\end{align*}

$h_{overlap}$ and $h_{union}$ represent the intersection and union in the height direction. The IoU loss is then defined as $L_{IoU} = 1 - IoU$.

\section{Autonomous Driving Datasets}
Increasing availability of labeled datasets with millions of samples allows training models more accurately. A brief summary of some of the most well-known datasets and competitions is presented in Table \ref{tbl:detection_datasets}.

\begin{table*}[t!]
  \caption{Autonomous Driving datasets that are used for 3D object detection.\label{tbl:detection_datasets} }
  \begin{tabular*}{\textwidth}{@{\extracolsep{\fill}}lccccccl}
    \hline
    Name& Year & PC frames &  RGB images & Classes & Diff. weather/time & Obj./frame &Location   \\
    \hline
    KITTI  \cite{GeigerVisionMeetsRobotics2013}          &    2012     &   15.4k     &    15k& 8  & No/No  &13& Karlsruhe (Germany)\\
    nuScenes\cite{caesarNuScenesMultimodalDataset2020}         &    2019    &   400k     &   1.4M  & 23 &  Yes/Yes &35    &Boston (USA), Singapore (SG)\\
    Waymo    \cite{sun2020scalability}           &    2019     &  200k     &   1M  &  4  &   Yes/Yes  &60 & SF , Phoenix, Mt. View (USA)\\
    ArgoVerse \cite{chang2019argoverse}        &    2019     &   44k    &   490k    &15&Yes/Yes& 45&Pittsburgh, Miami (USA)\\
    Lyft Lvl. 5\cite{houston2020one}    &    2020     &  323k     &    46k   &9   &   No/No &28&   Palo Alto (USA)\\
    A2D2    \cite{geyer2020a2d2}          &    2020    &   41k     &    41k   &   38&  Yes/No & - & Ingolstadt (Germany)\\
    LIBRE   \cite{carballo2020libre}       &   2020    &    -     &    -      & -  & Yes/Yes & - & Nagoya (Japan) \\
    ONCE \cite{mao2021one}        &  2021     &    16k      &    16k       & 5  & Yes/Yes  & 26 & China \\
    \hline
    IPS300+ \cite{wang2021ips300} &   2021    &    1.25k     &    2.5k      & 8  & No & 320 & China \\
    A9-Dataset \cite{cress2022a9}   & 2022      &    17.4k    &    68.4k     & 10 & Yes/Yes  & 34 & Munich (Germany)  \\
    \hline
  \end{tabular*}
\end{table*}

Datasets in autonomous driving contain multiple modalities to provide a robust and comprehensive understanding of the whole scene. Besides LiDAR information, most datasets also provide corresponding RGB images, which are dense and contain more semantic features to compensate for the sparsity in LiDAR point clouds. The cost to obtain RGB images is also much lower than getting point clouds. Some datasets also provide radar information such as \textit{nuScenes}. In comparison with the LiDAR sensor, radar has a farther scanning range and is less affected by extreme weather conditions. The first batch of the recently released \textit{A9-Dataset} contains infrastructure sensor data (camera and LiDAR) from the A9 Highway (including an accident) and a large traffic intersection with labeled vehicles and vulnerable road users (VRUs).
\section{Evaluation}

Table \ref{tbl:method_comparison} shows the \texttt{mAP} and inference speed of the mentioned \textit{state-of-the-art} methods (grouped by one-stage and two-stage detectors) on the \textit{KITTI}, \textit{nuScenes} and \textit{Waymo Level 1} test set.

\begin{table*}[t]
\renewcommand{\arraystretch}{1.3}
\caption{Comparison of the \textit{state-of-the-art} methods on the \textit{KITTI} (3D car, moderate), \textit{nuScenes} and \textit{Waymo Level 1} test set.}
\label{tbl:method_comparison}
\begin{tabular*}{\textwidth}{@{\extracolsep{\fill}}llllllll}
\hline
Method & &\multicolumn{2}{c}{KITTI} & \multicolumn{2}{c}{nuScenes} & \multicolumn{2}{c}{Waymo} \\
\hline
 & Year &3D mAP & FPS (ms) & NDS & FPS (ms) & 3D mAP & FPS (ms) \\
\hline
PointPillars \cite{langPointPillarsFastEncoders2019}& 2019 & 74.31\% & \textbf{62 (16 ms)}  & 0.44 & \textbf{62 (16 ms)} & 45.52\% & \textbf{62 (16 ms)} \\
3DSSD \cite{yang20203dssd}& 2020 &79.57\% & 26 (38ms) & - &  - &-&- \\
SA-SSD \cite{he2020structure}& 2020 &79.79\% & 25 (40ms) & - & - & 61.48\% & 25 (40ms) \\
CIA-SSD \cite{zheng2020cia} & 2021 & 80.28\% & 33 (30 ms) & - & - & - & -\\
SE-SSD \cite{zheng2021se} & 2021 & \textbf{83.73\%} & 33 (30 ms) & - & - & - & -\\
\hline
PV-RCNN \cite{shiPVRCNNPointVoxelFeature2020}& 2020 &81.43\% & 13 (80 ms) & - & - & 70.30\%& 4 (300 ms) \\
PV-RCNN++ \cite{shi2021pv}& 2021&\textbf{81.88\%} & 16 (60 ms)& - & - & 74.81\%& 10 (100 ms)\\
CenterPoint \cite{yin2021center}&2021& 74.87\% & \textbf{20 (51 ms)}&\textbf{0.71}& 16 (60 ms)& \textbf{79.25\%} &\textbf{11 (89 ms)}\\
\hline
 \end{tabular*}
\end{table*}

The \texttt{AP} on the \textit{KITTI} test set was evaluated on the moderate 3D object difficulty of the car class using 40 recall positions ($R40$). The rotated $IoU$ on the \textit{KITTI} test set was set to a threshold of $0.7$ for cars and $0.5$ for pedestrian and cyclists. \textit{PointPillars} runs at 62 FPS using \textit{TensorRT} for GPU acceleration. Using the \textit{PyTorch} pipeline, it runs at 42 FPS according to \cite{langPointPillarsFastEncoders2019}. Evaluation on the \textit{KITTI} test server was performed with a 1080TI GPU and an Intel i7 CPU (1 core @2.5 Ghz).
The overall 3D \texttt{mAP} on the \textit{Waymo Level 1} dataset was evaluated at an intersection-over-union (\texttt{IoU}) of 0.7. The detection region was set to 150 x 150 m ($[-75.2~m,~75.2~m]$ for the X and Y axis) and $[-2~m,~4~m]$ for the Z axis. According to \cite{shiExecutionSpeedPVRCNN2020} the inference speed of \textit{PV-RCNN} on the \textit{Waymo} dataset is about 300 ms (4 FPS).
\section{Challenges and Open Fields}


\label{cha:challenges_and_open_issues}

\textbf{Multi-frame 3D Object Detection.} Current state-of-the-art 3D object detection models mainly rely on LiDAR point clouds or its combination with other data types for reaching the best results. The dynamic nature of traffic causes occlusions that occur in fixed time intervals. LiDAR point cloud sparsity is another problem that hinders perfect 3D detection. The 3D object detection community has recently started to consider using data history with the release of large datasets such as \textit{nuScenes} \cite{caesarNuScenesMultimodalDataset2020} and \textit{Waymo Open Dataset} \cite{sun2020scalability} in addition to the \textit{KITTI Raw and Tracking} datasets \cite{GeigerVisionMeetsRobotics2013}. We introduce the most recent multi-frame 3D object detection studies in four primary categories with some obvious overlaps: (i) implicit multi-frame data processing, (ii) scene-level feature aggregation, (iii) object-level feature aggregation, and (iv) aggregation with attention mechanisms. 


We define implicit multi-frame approaches as not explicitly using learning-based methods for aggregating features from multiple frames. \textit{CenterPoint} \cite{yin2021center} estimates center offsets (velocities) between two successive frames for 3D detection and tracking. \cite{hu2020you} aggregates occupancy maps using Bayesian filtering through sequential point cloud sweeps. The method fuses the temporal occupancy map with the point cloud feature maps. In \cite{sun2021you}, object proposals are generated based on temporal occupancy maps, odometry data, and Kalman filter-based motion estimations. \cite{murhij2021real} aggregates two consecutive feature maps using odometry-based flow estimation. \cite{piergiovanni20214d} combines point clouds from multiple frames by tagging the points with timestamps, which are aligned using the ego-motion compensation and used as the input.  

The scene-level multi-frame feature map aggregation approach aims to construct a richer temporal feature map to enhance 3D object detection accuracy. \cite{luo2018fast} utilizes 3D convolutional layers to process multi-frame BEV voxel features across time for 3D detection, tracking, motion forecasting tasks. Instead of 3D convolutions, convolutional LSTMs are used for spatio-temporal BEV feature map aggregation in \cite{el2018yolo4d} and \cite{mccrae20203d}. Similarly, \cite{huang2020lstm} proposes a custom convolutional LSTM for multi-frame feature map fusion in addition to the ego-motion compensation for positional feature offsets. \textit{SDP-Net} \cite{zhang2020sdp} generates an aggregated feature map from sequential BEV features based on flow and ego-motion estimations. Detection and tracking results are obtained from the final temporal feature map. 

Object-level feature aggregation in time aims at obtaining better object representations using object feature vectors from preceding detections. \cite{ercelik2021tempfrustum} uses recurrent layers and 2D object tracking IDs to match and aggregate object features in time. \cite{yang20213d} generates multi-frame object proposal features from the object feature vectors in the memory using a non-local attention block for aligning features. \cite{qi2021offboard} processes LiDAR point cloud sequences for automatic labeling. A 3D multi-object tracker is used to match single-frame detections. A final stage refines the 3D bounding boxes after merging points of matched objects in time. \cite{xiong2021lidar} combines point clouds from multiple frames using the calibration information and generates proposals from the sizeable combined point clouds. A spatio-temporal GNN further refines the proposals. 

Multi-frame attention mechanisms have taken a growing interest in 3D object detection. \cite{yang20213d} applies non-local attention blocks to consecutive object features to align and generate temporal object features. \cite{yin2020lidar} and \cite{yin2021graph} use spatial and temporal transformer attention modules to align the temporal feature maps generated with a convolutional GRU. Similarly, \cite{yuan2021temporal} applies a spatio-temporal transformer module to obtain a temporally-enhanced feature map, which a detection head takes as input.

Multi-frame studies show that feature aggregation through successive frames enhances the 3D detection accuracy. However, alignment remains an issue. Methods so far have used ego-motion, calibration, or tracking-based alignment methods. Also, convolutional recurrent networks' filters expectedly account for such offsets between sequential feature maps. Attention mechanisms have stood out as a promising solution to the alignment issue. Even though many graph network-based 3D detection methods exist, direct feature alignment with graph networks for 3D detection is surprisingly a new open field. Therefore, transformer attention mechanisms and temporal graph representations remain prospective fields for multi-frame 3D detection.

\textbf{Imbalanced datasets and point cloud sparsity}. Class imbalance like in the \textit{KITTI} dataset (100k cars, 25k pedestrians, 7k cyclists) leads to a bad detection of pedestrians and cyclists. \textit{nuScenes} is about two times larger than the \textit{KITTI} dataset and contains more scenarios at different times (day, night) and weather conditions (sunny, rainy, cloudy). Each scene has a full 360 degree field-of-view. However, the number points per scene in \textit{nuScenes} is only a third of \textit{KITTI}. Detecting small objects in sparse and irregular LiDAR point clouds is difficult. The \textit{Waymo Open} dataset is one of the largest open source 3D object detection datasets so far in autonomous driving. One obvious drawback of the \textit{Waymo} dataset is, that it only provides 4 classes.\\

\textbf{Cross-dataset domain adaptation} \cite{wang2020train}. After training an algorithm on a large publicly available dataset, one often wants to apply this algorithm on own data. The problem here is often, that different sensors and mounting positions are used, so that the trained model is not able to detect objects. Car sizes for example are varying across different geographic regions. Most of the novel datasets contain cars that were only observed from one side. The depth of the bounding boxes must be estimated based on gathered experience during training.\\

\textbf{Data annotation}. Accurate data annotation in 3D space is a very tedious task and is often outsourced to expert annotators. Open source 3D annotation tools like \cite{zimmer3DBATSemiAutomatic2019} automate the annotation process so that a large number of annotations can be created in a short time. After annotating your own data and retraining your model on that annotated data, you should receive better detection and classification results. 
\section{Conclusion}
This paper has presented a survey of novel methods, \textit{state-of-the-art} 3D object detection architectures, and their comparison (in terms of mean average accuracy and inference speed). We hope this will help practitioners to choose an appropriate method when deploying object detection algorithms in the real world. We have also identified some new techniques for improving speed without sacrificing much accuracy, e.g. by using \textit{TensorRT} for GPU acceleration. The attention mechanism is recently being applied also in 3D object detection to further increase the \texttt{mAP}. In detail, we described novel data augmentation methods, sampling strategies, activation functions, attention mechanisms, and regularization methods. We listed recently introduced normalization methods, optimization methods, learning rate schedules, loss functions, and evaluation metrics. Finally, potential research directions, challenges and issues of 3D object detection methods in point clouds were given.

                                  
\section*{ACKNOWLEDGMENT}
This work was funded by the Federal Ministry of Transport and Digital Infrastructure, Germany as part of the Providentia++ research project (Grant Number: 01MM19008A). The authors would like to express their gratitude to the funding agency and to the numerous students at TUM for technical editing, language editing, and proofreading. 

\bibliographystyle{IEEEtran}


\end{document}